# Exploiting Neighborhood Structural Features for Change Detection

Abstract: In this letter, a novel method for change detection is proposed using neighborhood structure correlation. Because structure features are insensitive to the intensity differences between bi-temporal images, we perform the correlation analysis on structure features rather than intensity information. First, we extract the structure feature maps by using multi-orientated gradient information. Then, the structure feature maps are used to obtain the Neighborhood Structural Correlation Image (NSCI), which can represent the context structure information. In addition, we introduce a measure named matching error which can be used to improve neighborhood information. Subsequently, a change detection model based on the random forest is constructed. The NSCI feature and matching error are used as the model inputs for training and prediction. Finally, the decision tree voting is used to produce the change detection result. To evaluate the performance of the proposed method, it was compared with three state-of-the-art change detection methods. The experimental results on two datasets demonstrated the effectiveness and robustness of the proposed method.

## 1. Introduction

Change detection refers to the use of bi-temporal remote sensing images acquired from the same geographical area but at different times to extract and analyze the ground change information (Shafique et al. 2022). With the advancement of remote sensing technology, change detection in remote sensing images becomes more and more important, and it has attracted widespread interest due to it being extensively used in many remote sensing applications (Jiang et al. 2022; Ye et al. 2022), such as urban planning (Ding et al. 2015), environment monitoring (Yetgin 2012), and nature disaster assessment (Byun et al. 2015), (Hao et al., 2018), etc.

During the past several decades, a large number of change detection methods have been proposed in the remote sensing community. Based on the granularity of the adopted basic processing unit, the existing change detection methods can be divided into pixel-based change detection (PBCD) and object-based change detection (OBCD) (Jiang et al. 2022). PBCD methods only take the individual pixel as its basic unit without considering any contextual and geometric information among the adjacent pixels. Although the PBCD methods are simple and easy to operate and have been widely used, the accuracy is often degraded due to image noise (Peng et al. 2021). The OBCD methods use the "object" as the analysis unit, and the "object" is a group of adjacent local pixel clusters that are obtained through image segmentation using spectral texture features (Lei et al. 2022), geometric features (such as shape and contour). However, due to the limitations of extracting image objects by segmenting bi-temporal images, the "object" is prone to over-segmentation and boundary without semantic integrity (Zhang et al., 2021). In recent years, with the rapid development of deep learning (DL) technology, DL-based change detection has become a hot topic, and a growing number of researchers have used DL techniques for change detection (Mou et al. 2019; Ren et al. 2021). DL-based methods can automatically extract semantic and discriminative features from bi-temporal images. However, the DL-based methods are known to require large amounts of labeled data for proper training, which is a limitation because producing a huge amount of labeled data is a time-consuming and tedious task.

Many studies have been conducted to improve the accuracy of change detection results by using texture and spatial neighborhood information (Lv et al. 2018). For example, Li et al. (2017) proposed a change detection method by using texture and spectral information for remote sensing images to improve the accuracy. Im et al. (2005) proposed a change detection model based on the Neighborhood Correlation Image (NCI).However, these methods directly use intensity of pixels to extract neighborhood context information. As a result, they are not optimal in the change detection of remote sensing images because there are usually significant intensity and spectral differences between bi-temporal images. Compared with intensity information, spatial structure features are preserved between bi-temporal images in despite of intensity differences (Ye et al. 2017). Accordingly, we attempt to exploit structure features of local neighborhood for change detection.

In order to overcome the limitation of the intensity differences between bi-temporal images, this paper proposes a novel change detection method based on neighborhood structure correlation. The main contributions of this paper are summarized as follow:

(1) A novel change detection method is proposed for remote sensing image change detection. Unlike existing works, we exploit the structural features of local neighborhood from bi-temporal images for change detection.

(2) A more robust neighborhood feature named Neighborhood Structural Correlation Image (NSCI) is proposed, which are calculated by using structure features rather than raw intensity information. Moreover, we also propose a novel measure named matching error (ME), which can reflect the spatial neighborhood correlation between bi-temporal images.

## 2. Implementation

In this section, we will give the details of the proposed method. As shown in Figure 1, the bi-temporal structure feature maps are first extracted from image $T_1$ and $T_2$, respectively. Then, the NSCI and ME are calculated by using the structure feature maps. Subsequently, a feature vector with 4 dimensions is generated by combining the NSCI (i.e., correlation, slope, and intercept) and ME features for each pixel, and a change detection model based on RF is constructed, more details are descripted as follow.

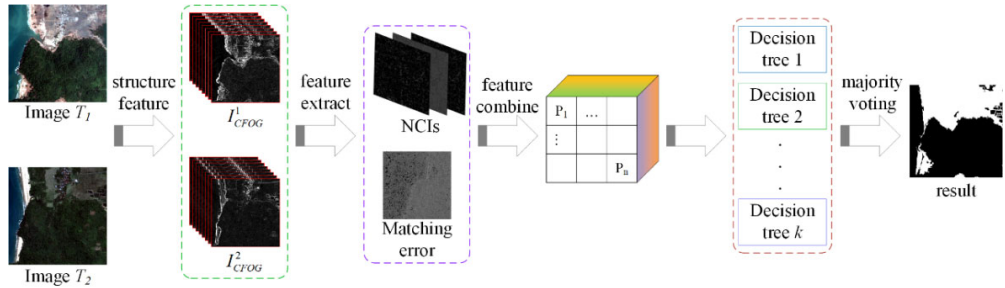

Figure 1. Flowchart of the proposed method.

### 2.1. Extracting Geometric Structure Feature

Considering that intensity differences are common between bi-temporal images, structure features are used to extract neighborhood information because it is more robust to spectral (intensity) differences caused by time changes and atmospheric conditions. In this letter, we extract structure features by using a robust geometric feature descriptor named channel feature of orientated gradients (CFOG) (Ye et al., 2019).

Figure 2 shows the extraction process of CFOG, which mainly includes the two following steps. (1) Constructing Orientated Gradient Channel: we first compute the multi-orientated gradient features of an image, then these gradient features are arranged in the Z-direction to generate the orientated gradient. (2) Constructing Structural Features: a convolved feature channel is first obtained by a 2-D Gaussian kernel in X- and Y- orientation and a kernel $[1,2,1]^T$ in the Z-orientation. Then a convolved feature channel is normalized to form the final CFOG features.

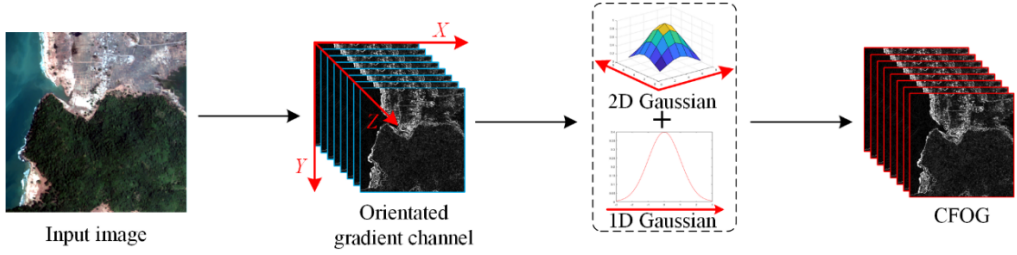

Figure 2. Construction process of CFOG.

## *2.2. Neighborhood Information Feature*

Neighborhood information is one of the commonly used features in change detection (Im and Jensen, 2005). However, the traditional NCI is directly calculated by using raw intensity information of bi-temporal images, which is vulnerable to intensity and spectral differences between bi-temporal images. Considering that geometric structure features are insensitive to intensity differences, this paper propose a more robust neighborhood feature (i.e., NSCI), which are calculated by using the CFOG structure feature rather than raw intensity of bi-temporal images. Similarity, the NSCI between two bi-temporal CFOG features within a specified neighborhood is calculated using the following equations:

$$\text{cov}_{12} = \frac{\sum_{i=1}^{n}(BV_{i1}-\mu_1)(BV_{i2}-\mu_2)}{n-1} \quad (1)$$

$$r = \frac{\text{cov}_{12}}{s_1 s_2} \quad (2)$$

$$a = \frac{\text{cov}_{12}}{s_1^2} \quad (3)$$

$$b = \frac{\sum_{i=1}^{n} BV_{i2} - a \sum_{i=1}^{n} BV_{i1}}{n} \quad (4)$$

where $r$ is Pearson's product-moment correlation coefficient, $\text{cov}_{12}$ is the covariance between feature values found in all layers of two bi-temporal CFOG feature images in the neighborhood, $s_1$ and $s_2$ are the standard deviations of feature values found in all layers of the bi-temporal images in the neighborhood, respectively. $BV_{i1}$ and $BV_{i2}$ are the *i*th feature values of the pixels found in all layers of the CFOG feature image $T_1$ (named $I_{CFOG}^1$) and the CFOG feature image $T_2$ (named $I_{CFOG}^2$), respectively. And $n$ is the total number of the pixels found in all layers of each structure images within the neighborhood. $\mu_1$ and $\mu_2$ are the means of feature values found in all layers of images $I_{CFOG}^1$ and $I_{CFOG}^2$ within the neighborhood, respectively. Slope $a$ and Intercept $b$ are calculated by using least square estimates.

## *2.3. Matching Error Feature*

Although NSCIs can reflect the spatial neighborhood feature information between bi-temporal images, it is calculated by using a specified neighborhood window size. If the neighborhood window size is small, some noise will be introduced in the result of change detection. And if larger neighborhood window size is used for removing this noise, some linear feature changes are removed. In order to make full use of the spatial neighborhood information between bi-temporal images, this letter introduces the matching error (ME) to enhance the representation of spatial neighborhood information. ME is measured by using the correlation coefficient of local neighborhood between structure feature maps with a template matching manner. For the bi-temporal images that have been aligned exactly, the neighborhood of unchanged area centered on pixel $P_{unchange}$ tends to have higher correlation, and the position of the neighborhood with the highest correlation has almost no translation or the translation is quite small (i.e., the ME is quite small). Whereas, the neighborhood of change area centered on pixel $P_{change}$ with the highest correlation will be far away from the location of neighborhood (i.e., the ME is quite large).

Let us give a formal definition for the ME. For any pixel $P$ of image $T_1$, a template with a specific size ($3\times 3$ pixels) centered on $P$ is selected. Then, the search region with a larger size ($9\times 9$ pixels) is selected on image $T_2$. To calculate the ME of the pixel $P$, the template slides in the search region to compute the correlation coefficient. Such operation will yield a group of matrices which has the same row and column to the search window. Finally, the matching error is obtained by calculating the Euclidean distance between the pixel point with the maximum correlation coefficient and the center of the matrices. The ME of $P$ is defined as:

$$ME_{max} = \arg\max\left(\frac{\sum_{x,y}(F(x,y)-\bar{F}_{u,v})(G(x-u,y-v)-\bar{G})}{\sqrt{\sum_{x,y}(F(x,y)-\bar{F}_{u,v})^2 \sum_{x,y}(G(x-u,y-v)-\bar{G})^2}}\right) \quad (5)$$

$$ME_{val} = \sqrt{(P_x - ME_x)^2 + (P_y - ME_y)^2} \quad (6)$$

where $u$ and $v$ are variable shift component along x-direction and y-direction, respectively. $F$ and $G$ are the search region and template of image $T_1$ and $T_2$, respectively. $\bar{F}_{u,v}$ denotes the mean value of $F(x,y)$ within the area of template $G$ shitted to $(u,v)$, and $\bar{G}$ is the mean value of the template $G$. $ME_{max}$ denotes the position of the largest correlation coefficient between the template and the search region. $ME_x$ and $ME_y$ are the x-position and y-position of $ME_{max}$. $P_x$ and $P_y$ are the centered position of the search region $F$, respectively. $ME_{val}$ is the ME.

As shown in Figure 3, the unchanged pixel and changed pixel are labeled with blue cross hairs and red cross hairs, respectively. The template and search region is marked by using solid rectangle and dashed rectangle, respectively. Figure 3(c)-(d) illustrate the ME matrices of unchanged pixel and changed pixel, respectively. In Figure 3(c), it can be observed that the position of the pixel of maximum correlation coefficient is at the center of matrices, which is considered as the unchanged pixel. Whereas, it is obvious that the position of the pixel with maximum correlation coefficient is far from the center of the matrices [Figure 3(d)], which is considered as the changed pixel.

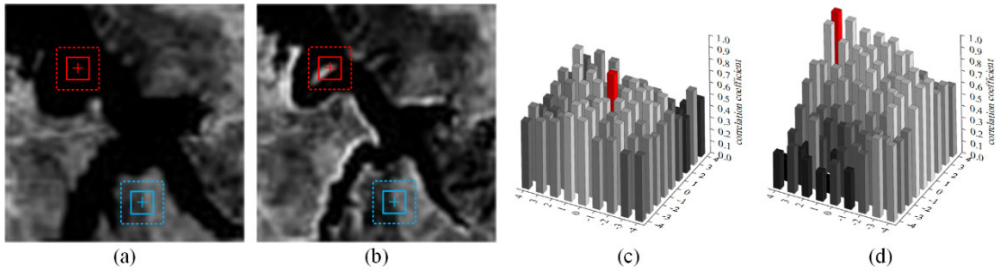

Figure 3. Matrices of Matching Error. (a) Image $T_1$. (b) Image $T_2$. (c) Matching Error Matrices of Unchanged Pixel. (d) Matching Error Matrices of Changed Pixel.

## *2.4. RF-Based Changed Detection Strategy*

In general, the change detection can be considered as a two-class classification issue. Accordingly, a change detection model based on RF is constructed in this letter. RF is an ensemble classification method that evolves from the bagging method proposed by Breiman (Breiman 2001). Generally, growing an RF model involves three processes. Firstly, using the bagging approach, $k$ subsets of training data are created by randomly sampling with replacement, where $k$ is the number of trees in the RF model. After that, a sample of the feature subspace is taken from the original feature space (i.e. only some of all the features are used to build tree classifiers.). Finally, integrating the $k$ classification trees to form a random forest and the final classification decision is determined by a majority vote of all classification trees.

$$H(x) = \arg\max_{y_i} \sum_{i\in[1,2,...,k]} I(h_i(x)=y_j), j=1,2,...,C \quad (7)$$

where $H(x)$ is the combination model, $h_i(x), i=1,2,...,k$ is the decision tree model according to $k$ subsets of training data, $y_j, j=1,2,...,C$ is the labels of the $C$ classes, where its values is 2. And $I(\cdot)$ is the combined strategy, which is defined as:

$$I(x) = \begin{cases} 0, h_i(x) = y_j \\ 1, h_i(x) \neq y_j \end{cases} \quad (8)$$

where $h_i(x)$ is the output of the decision tree, $y_j, j=1,2$ is the class label of the label classes. Based on the RF model, the combined Neighborhood features (i.e., NSCI and ME) are adopted to perform change detection. Firstly, for every pixel, a feature vector with four dimension is produced. Then, the training set is used to build the RF model. Finally, the change detection result is generated by assembling all specific trees of the RF model.

## 3. Experimental and Results

In order to verify the effectiveness of the proposed method, three change detection methods are tested and compared: The Change Vector Analysis (CVA) (Polykretis et al., 2020), the NCI (Im and Jensen, 2005), and the Adaptive Contextual Information (ACI) (Lv et al., 2018). Furthermore, we design the ablation experiments to verify the validity of the proposed ME metric. Table I shows the attribute characteristics of the five methods. We use four evaluation metrics to quantitatively evaluate the performance of experiment results, which are overall accuracy (OA), false alarms (FAs), missed detections (MDs) and the Kappa coefficient (KC).

Table 1. Change Detection Methods Involved in This letter.

| Attribute \ Method | CVA | NCI | ACI | NSCI | NSCI+ME(proposed) |
|---|---|---|---|---|---|
| Intensity information | ✓ | ✓ | × | × | × |
| Neighborhood information | × | ✓ | ✓ | × | ✓ |
| Structure feature | × | × | × | ✓ | ✓ |

### 3.1. Description of the Experiment Datasets

The first dataset is a pair of QuickBird images with the size of 1500 × 1500 pixels, four bands, and a spatial resolution of 2.5m. And the two images locate in Indonesia, which were acquired in April 2004 and Jan 2005, respectively. The images was classified into three major classes—water, vegetation, and bare area, which were representative to describe the changes during the period. The second dataset is a pair of multitemporal Landsat-5 images with the size of 300×412 pixels in Sardinia Island, which was acquired in September 1995 and July 1996, respectively. The two datasets in $T_1$ image and $T_2$ image as shown in Figure 4(a) and (b), and Figure 4 (c) shows the ground truth.

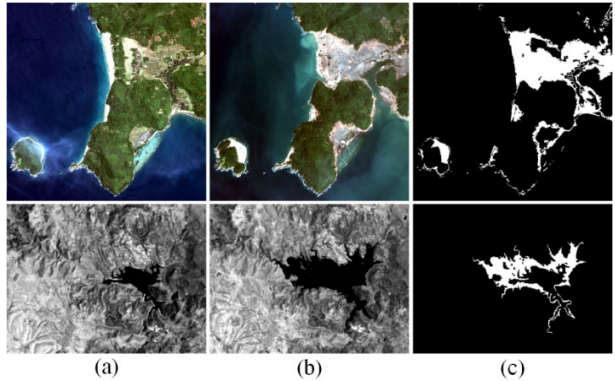

Figure 4. Illustration of two datasets. (a) Image $T_1$. (b) Image $T_2$. (c) Ground truth. (The first and second rows are the first dataset and the second dataset, respectively.)

## 3.2. Results and Analysis

The quantitative evaluation results of different methods are showed in Table II. for the Indonesia dataset, NCI obtains the lowest OA and KC of 90.72% and 0.65 among these methods. The followed by CVA which obtain an OA of 91.15% and KC of 0.67, respectively. the OA and KC of the proposed NSCI is improved by 5.07% and 0.14 compared with NCI, respectively, which denotes that structure feature information is more robust than intensity information. The OA and KC of the proposed NSCI+ME was improved by 1.06% and 0.04 compared with NSCI, respectively, which demonstrates that the ME metric can enhance the representation of neighborhood correlation information and improve the accuracy. For the Sardinia dataset, the OA of the proposed NSCI+ME method is improved by 2.95%, 2.81%, 2.12%, and 0.37% compared with CVA, NCI, ACI, NSCI, respectively. And the KC is improved by 0.16, 0.13, 0.12, and 0.03, respectively. These experimental results indicate that the proposed method is superior to other methods in performance.

Table 2. Accuracy Comparison of Different Methods over Two Dataset. The Best Score is Marked in Bold.

| Datasets | Methods | OA(%) | FAs(%) | MDs(%) | KC |
|---|---|---|---|---|---|
| | CVA | 91.15 | 37.23 | 13.84 | 0.67 |
| | NCI | 90.72 | 42.08 | **10.90** | 0.65 |
| Indonesia | ACI | 94.84 | 12.59 | 25.15 | 0.75 |
| | NSCI | 95.79 | 12.96 | 15.75 | 0.79 |
| | NSCI+ME | **96.85** | **10.81** | 14.89 | **0.83** |
| | CVA | 90.06 | 4.01 | **9.33** | 0.69 |
| | NCI | 90.20 | 3.48 | 18.10 | 0.72 |
| Sardinia | ACI | 90.89 | 3.12 | 10.08 | 0.73 |
| | NSCI | 92.64 | 1.96 | 13.50 | 0.82 |
| | NSCI+ME | **93.01** | **1.93** | 12.96 | **0.85** |

Figure 5 provides a more intuitive visualization of each method's result on the two datasets. In Figure 5 (b), many "salt and pepper" noisy spots exist in the results of CVA. In addition, for the Indonesia, CVA cannot correctly identify the difference between pixels in the vegetation area, the reason is that the image intensity information is sensitive to spectral differences. In the result of NCI, for Indonesia dataset [Figure 5(c)], the change detection result has achieved improvement compared with CVA, the reason is that NCI adopts neighborhood information within the specified window size. Figure 5(d) shows the results of the ACI method. The small "salt and pepper" noisy spots were eliminated by using the adaptive neighborhood contextual region. However, ACI takes the intensity mean value of the extended region as the central pixel value, which will affect the accuracy of the boundary of the changed areas. In Figure 5(f), it can be observed that compared with NSCI, the proposed method yields more accuracy changed regions. In addition, Figure 5(e)-(f) present the ablation results on the two datasets, the result of NSCI+ME retain the boundary more completely [Figure 5(f)] compared with that of only using NSCI [Figure 5(e)]. The reason is that the proposed NSCI+ME method not only effectively made use of the structure information between bi-temporal images, but also taking the spatial neighborhood information between bi-temporal images into account.

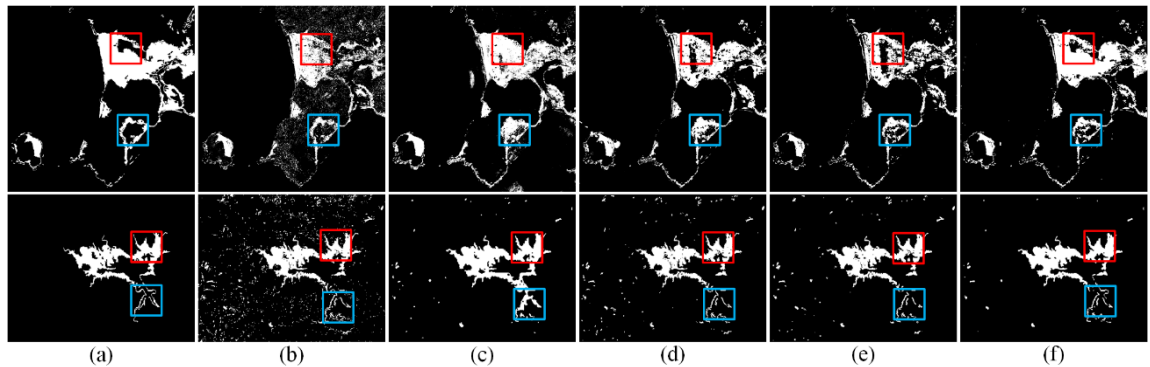

Figure 5. Visualized results by different methods over two datasets (a) Ground truth. (b) CVA. (c) NCI. (d) ACI. (e) NSCI. (f) NSCI+ME.

## 4. Conclusion

In this letter, a novel change detection method is proposed by combining geometric structure feature and neighborhood information. Considering that the traditional NCI method was limited in using intensity of pixels, we build a novel NCI feature (named NSCI) using neighborhood structure information that is more robust to spectral differences than intensity information. Furthermore, we also propose an additional measure (named ME), which adopts the matching error to further depict the neighborhood correlation between bi-temporal images. Then, NSCI and ME are used as the model inputs for training and prediction. Finally, the change detection result is generated by the constructed RF model. The experiments were carried out on two datasets to confirm the effectiveness of the proposed method.

**Declaration of Competing Interest**

The authors declare that they have no known competing financial interests or personal relationships that could have appeared to influence the work reported in this paper.
**Funding**

This research is supported by the National Natural Science Foundation of China (No.41971281).